 \documentclass[pmlr,twocolumn,10pt]{jmlr} 

\usepackage{booktabs}
\usepackage{siunitx}
\usepackage[switch]{lineno}

\usepackage{graphicx}
\usepackage[utf8]{inputenc} 
\usepackage[T1]{fontenc}    
\usepackage{hyperref}       
\usepackage{url}            
\usepackage{booktabs}       
\usepackage{amsfonts}       
\usepackage{nicefrac}       
\usepackage{xcolor}         
\usepackage{amsmath,amssymb,amsfonts}
\usepackage{soul}
\usepackage{xcolor}
\usepackage{wrapfig,lipsum,booktabs}
\usepackage{natbib}
\usepackage{url}
\usepackage{xspace}
\usepackage{multirow}
\usepackage{booktabs}
\usepackage{stfloats}
\usepackage{graphicx}
\usepackage{subcaption}
\usepackage{array}
\usepackage{colortbl}
\usepackage{listings}
\definecolor{lightgray}{gray}{0.85}
\usepackage{comment}
\usepackage[T1]{fontenc}
\usepackage{threeparttable}         
\usepackage{algorithm}
\usepackage{algorithmicx}
\usepackage{algpseudocode}
\usepackage{changepage}

\definecolor{codegreen}{rgb}{0.25,0.5,0.5}
\definecolor{codeblue}{rgb}{0.0,0.5,0.7}

\lstdefinestyle{myMatlab}{
language=Matlab,
keywordstyle=\color{javaLila},
commentstyle=\color{javaGreen},
stringstyle=\color{javaGreen},
numbers=left,
stepnumber=1,
numbersep=5pt,
numberstyle=\tiny,
breaklines=true,
breakautoindent=true,
breakatwhitespace=false,
postbreak=\space,
tabsize=2,
basicstyle=\ttfamily\scriptsize,
showspaces=false,
showstringspaces=false,
extendedchars=true,
backgroundcolor=\color{white},
 morekeywords={classdef}
 }

\newcommand{\equal}[1]{{\hypersetup{linkcolor=black}\thanks{#1}}}

\theorembodyfont{\upshape}
\theoremheaderfont{\scshape}
\theorempostheader{:}
\theoremsep{\newline}
\theoremsep{\newline}

\jmlrvolume{LEAVE UNSET}
\jmlryear{2024}
\jmlrsubmitted{LEAVE UNSET}
\jmlrpublished{LEAVE UNSET}
\jmlrworkshop{Machine Learning for Health (ML4H) 2024} 

\newcommand{\method}{\textbf{BernGraph}\xspace}

\title{
BernGraph: Probabilistic Graph Neural Networks for EHR-based Medication Recommendations
}

\author{%
\Name{Xihao Piao}\equal{Equal contribution} \Email{xihao.piao@sanken.osaka-u.ac.jp}\\
\addr SANKEN, Osaka University, Japan
\AND
\Name{Pei Gao}\footnotemark[1] \Email{pei.gao@naist.jp}\\
\addr Nara Institute of Science and Technology, Japan
\AND
\Name{Zheng Chen} \Email{zheng.chen@sanken.osaka-u.ac.jp}\\
\addr SANKEN, Osaka University, Japan
\AND
\Name{Lingwei Zhu} \Email{lingwei.zhu@ualberta.ca}\\
\addr University of Alberta, Canada
\AND
\Name{Yasuko Matsubara} \Email{yasuko.matsubara@sanken.osaka-u.ac.jp}\\
\addr SANKEN, Osaka University, Japan
\AND
\Name{Yasushi Sakurai} \Email{yasushi.sakurai@sanken.osaka-u.ac.jp}\\
\addr SANKEN, Osaka University, Japan
\AND
\Name{Jimeng Sun} \Email{jimeng@illinois.edu}\\
\addr University of Illinois Urbana-Champaign, USA
}

\begin{document}
\maketitle

\begin{abstract}
Electronic Health Records (EHRs) frequently contain binary event sequences, such as diagnosis, procedure, and medication histories. These sequences, however, are characterized by a large percentage of zero values, presenting significant challenges for model learning due to the scarcity of gradient updates during processing.
We propose a novel approach that reframes EHR data from a statistical perspective, treating it as a sample drawn from patient cohorts and transforming binary events into continuous Bernoulli probabilities. This transformation not only models deterministic binary events as distributions but also captures event-event relationships through conditional probabilities.
Building upon this probabilistic foundation, we implement a graph neural network that learns to capture complex event-event correlations while emphasizing patient-specific features. Our framework is evaluated on the task of medication recommendation, demonstrating state-of-the-art performance on large-scale databases.
Extensive experimental results show that our proposed \method\footnote{The source code is available at \url{https://anonymous.4open.science/r/BEHRMecom-031C}} achieves 0.589, 0.846, and 0.844 of Jaccard, F1 score, and PRAUC, respectively, on the MIMIC-III dataset, with improvements of 4.6\%, 6.7\%, and 6.3\% compared to baselines, including those using secondary information. 
This work presents a promising direction for handling sparse binary sequences in EHRs, potentially improving various healthcare prediction tasks.
   
   
\end{abstract}

\section{Introduction}

\begin{figure}[t]
\centering
\includegraphics[width=\columnwidth]{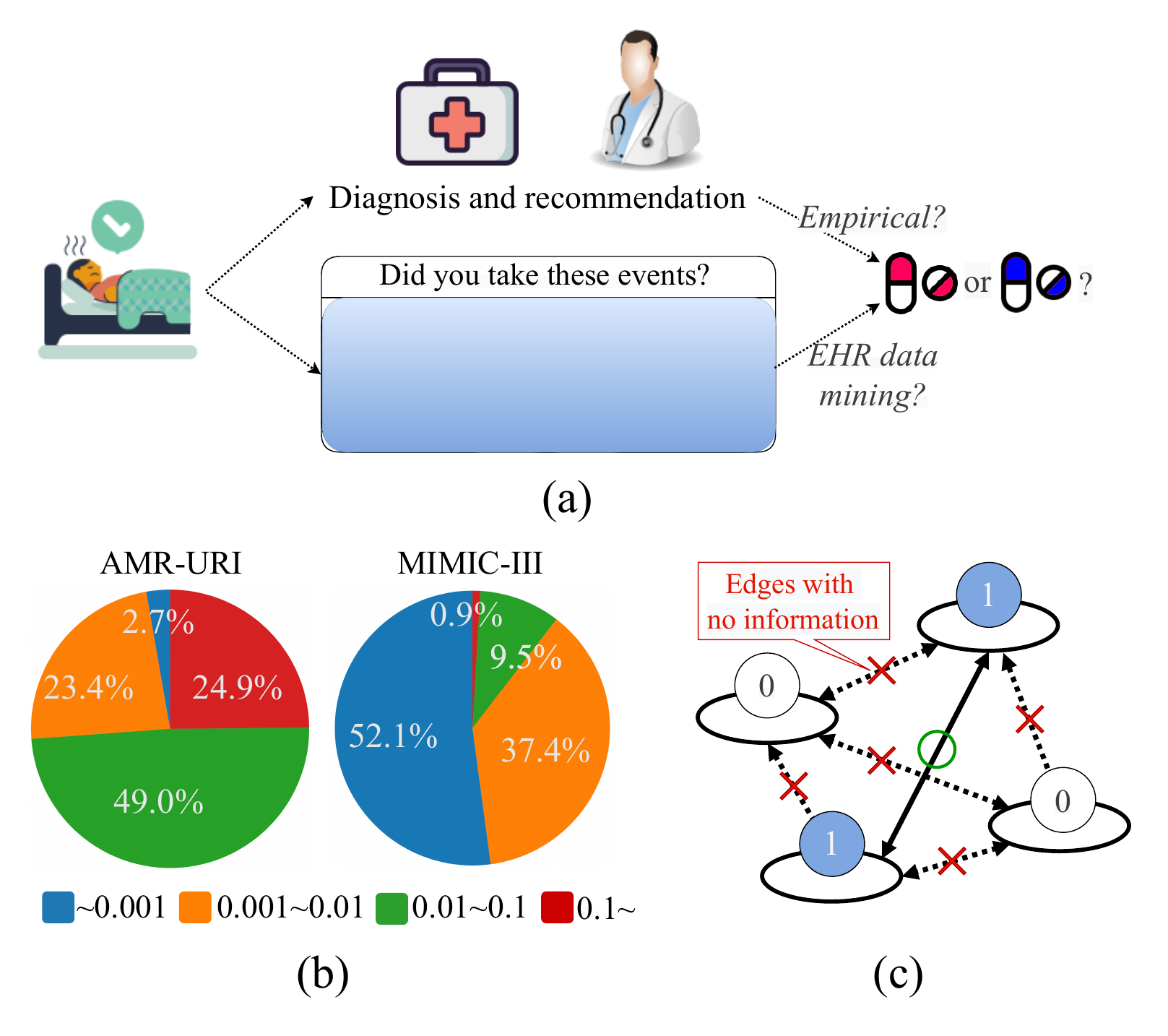}
\caption{
(a) Medication recommendations based on historical medical events. (b) The vast majority of event data are zero values. 
For example, the proportion of non-zero values exceeding 10\% is lower than 0.9\% in MIMIC-III.
(c) Those zero values do not propagate learning signals.
}
\label{fig:ehrproblem}
\end{figure}

Medication errors are one of the most serious medical errors that could threaten patients' lives.
Studies indicate that over 42\% of these errors stem from physicians or doctors with insufficient experience or knowledge concerning specific drugs and diseases \citep{review42issue}.
This challenge becomes even more pronounced with more new drugs and treatment guidelines.
On the other hand, many patients are concurrently diagnosed with multiple diseases, complicating the decision-making process for doctors when determining more appropriate remedies for patients \citep{wwwcognet}.
To this end, the medical community will benefit from an automated medication recommendation system that can assist doctors in decision-making.
In the past decades, a vast amount of clinical data representing patient health status (e.g., medical reports, radiology images, and allergies) has been collected.
This has remarkably increased digital information, known as electronic health records (EHR), available for patient-oriented decision-making.
In this context, mining EHR data for medication recommendation has attracted growing research interests \citep{wwwcognet}, as illustrated in Fig. \ref{fig:ehrproblem}(a).

Successful methods typically rely on modeling outcomes of various medical events and diagnoses by observing longitudinal patient histories, i.e., tracking state variations across hospital visits \citep{retain,2017-LEAP,safedrug,wwwcognet}.
They often employ sequential neural networks, such as RNNs or Transformers.
They construct a feature space centered around visits, each characterized by the current state of various medical events.
Their objective is to discern long-range temporal dependencies that reflect the historical progression of diseases. 
These methods, however, suffer from a number of drawbacks.
From a data viewpoint, different EHR databases adopt distinct visit recording methodologies \citep{singh2015incorporating}. 
For example, the AMR-URI database \citep{Michael2020AMR-UTI} employs a regular visit record, while visits in the MIMIC database \citep{Johnson2016MIMICIIIAF} are patient-specific.
Even within the same database, the time intervals between visits can differ greatly between patients.
From a model perspective, a common obstacle is the "cold-start" issue, where medication recommendation systems struggle due to insufficient patient-specific historical data \citep{ye2021unified}.
In practice, this issue becomes particularly pronounced when the system encounters newly onboarded patients with scarce or even absent historical data (i.e., visits) about their medical backgrounds.
While many researchers treat irregular time intervals as input variables \citep{nguyen2016mathtt,miotto2016deep,xiang2019time} or apply the carry-forward imputation method \citep{luo2022evaluating}, these approaches may introduce biases to temporal dependencies.

Some studies focus solely on current diagnoses of patients and medical event records, modeling event-event correlations for recommendation purposes without considering historical visits \citep{2017-LEAP, 2020gate}.
However, a performance gap remains when compared to visit-based methods.
A typical medical event has a binary value indicating its outcome: 1 for occurrence and 0 for non-occurrence, which leads to a sparsity issue in EHR data.
The sparsity, characterized by a high proportion of zero entries in event sequences, can significantly impact the performance of recommendations (seen in Fig. \ref{fig:ehrproblem}(b)).
Zero values provide no tangible information for deep learning models, preventing the activation of weight propagation.
This often leads to a vanishing gradient problem and results in collapsed representations for medication recommendations (as shown in Fig. \ref{fig:ehrproblem}(c)). 
Previous studies employ secondary information, e.g., Drug-Drug Interaction Knowledge Graph (DDI-KG), to complement information to medical events \citep{wwwcognet}.
Yet, sparsity remains a persistent issue.

Therefore, this paper presents \method, a binary EHR data-oriented drug recommendation system.
Our method is built upon a graph neural network (GNN) and solely utilizes the EHR binary medical events for recommendation.
To tackle the sparsity in binary event sequences, we take a statistical perspective that allows us to transform the $(0, 1)$ values into continuous probabilistic representations and estimate them from patient cohorts.
This strategy has demonstrated substantial efficacy in modeling graphical structures and correlations within binary EHR data.
Architecturally, diverse events are incorporated as node attributes within the GNN.
When two events co-occur, we model their relationship as conditional Bernoulli probabilities to initialize the edge between them. 
This allows GNN to learn the correlations between events across different patients for medication recommendations.
Our results demonstrate that our method outperforms several benchmarks, including the latest state-of-the-art (SOTA) visit-based methods.
Specifically, our method shows an improvement on the MIMIC III dataset of approximately 4.8\%, 6.7\%, 6.3\%, and 4\% in terms of Jaccard, F1-score, PRAUC, and AUROC, respectively.
Moreover, \method is a simple and effective model that complements existing research in which the EHR data is often exploited in conjunction with secondary data but without transformations.

\section{Related Works}\label{sec:related}
\noindent\textbf{- EHR-based Medication Recommendation.}
\noindent\emph{Historical Visit-based Methods:}
Many studies leverage deep learning to extract features from EHR visit variations \citep{MLHC-2016-doctorAI, 2018-dynamictreat-RNN, 2019-orderfree-RNN, MLHC-2021-pointprocess, 2020gate}. 
For example, DoctorAI \citep{MLHC-2016-doctorAI} employs RNNs to capture dependencies between patient histories and doctor recommendations. MICRON \citep{changematter} focuses on medication changes and combinations from the last visit. 
Some works \citep{safedrug, wwwcognet} further incorporate additional information like Drug-Drug Interactions (DDI) to aid the model predictions.\\
\noindent\emph{Current Medical Event-based Methods:}
On the other hand, some works aim to analyze the patient's current health status: recommendations are made based on the EHR records from each patient visit \citep{2017-LEAP, gong2021smr, 2023-adma}.
LEAP \citep{2017-LEAP} encodes patient information in a one-hot encoding and then learns the relationship between patient features and drugs. 
SMR \citep{gong2021smr} combines EHR data with an external medical knowledge graph to leverage external medical expertise in decision-making.
MT-GIN \citep{2023-adma} models event-event correlations in EHR data using a graph-based approach, with a focus on events that co-occur. 
This approach overlooks the information that 0 values in EHR binary data can potentially provide.

\noindent\textbf{- Graph Modeling for Medication Recommendation.}\label{subsec:gnn}
Recent works leverage graph structures to model event correlations \citep{MLHC-2016-doctorAI, 2020learningStructure, 2020gate, MLHC-2023-Wharrie_Yang_Ganna_Kaski_2023,gamenet, gong2021smr, 2022-safedrug, wwwcognet, 2023-adma}. Gamenet \citep{gamenet} and Cognet \citep{wwwcognet} use GNNs to capture correlations, integrating DDI data for representation learning. GATE \citep{2020gate} proposes a GNN-based approach using events from a single visit. COGNet \citep{wwwcognet} further integrates both EHR and DDI graphs into a generative model. However, these approaches are designed based on binary medical event data.

\noindent\textbf{- Binary Data Representation.}\label{subsec:binary}
Mining binary data is challenging due to the varied interpretations of 0s and 1s across contexts \citep{1999-BinaryDataIssue, 2017-KDD-BinaryCluystering}. Additionally, 0s often contribute little to deep learning. In the EHR context, medGAN and POPCORN transform binary events into continuous probabilities \citep{2017-PMLR-medGAN, MLHC-2021-pointprocess}, but these focus on historical data or overall distributions. Developing an effective method to represent binary events directly in EHR data remains challenging.

\begin{figure*}[t]
    \centering
 	\includegraphics[width=2\columnwidth]{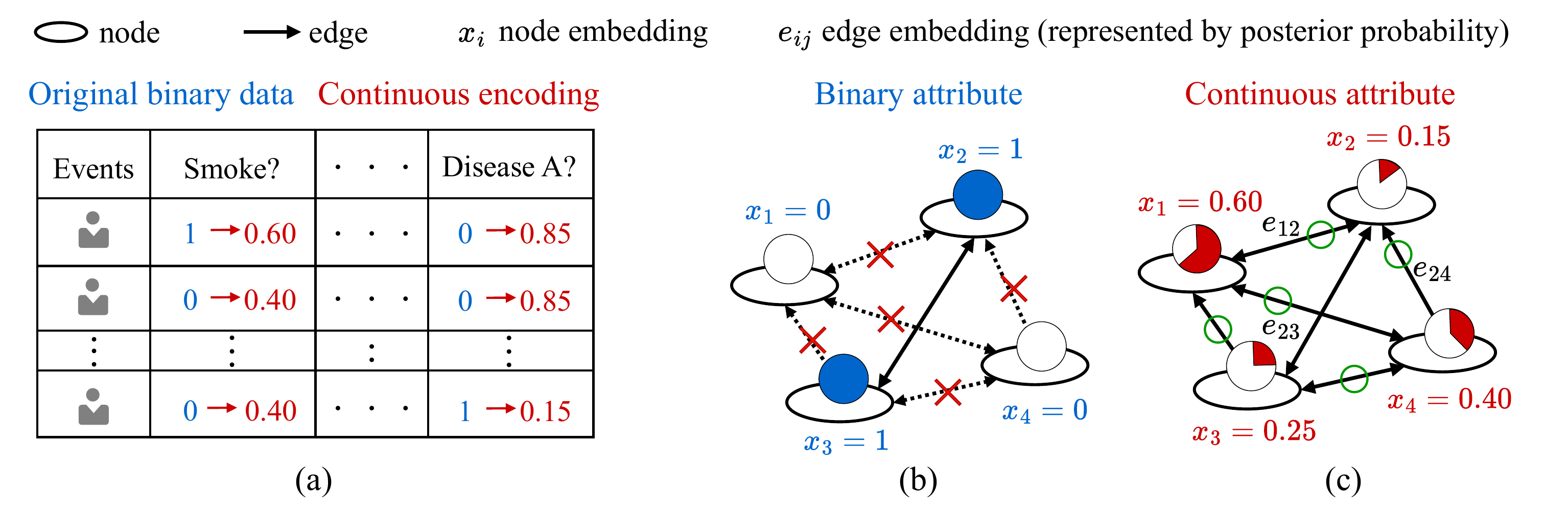}
 	\caption{
  (a) We transform $\{0, 1\}$ EHR entries into Bernoulli means. 
  (b) Learning directly from $\{0, 1\}$ data hinders signal propagation. 
  (c) Our method learns from continuous entries, with edges constructed from Bernoulli means and nodes from conditional Bernoulli probabilities (see Algorithm \ref{alg:generate_graph}).}
    \label{fig:main_graph}
 \end{figure*}

\section{Problem Formulation}\label{sec:prob}
\subsection{Recommendation System Based on EHR}

EHR data consists of various medical events, such as multiple symptoms, recorded chronically from patients. 
Each data sample contains binary outcomes of these events.
Let $\boldsymbol{x} \in \{0, 1\}^M$ denote a patient EHR sample, where $M$ represents the total number of medical events, $1$ stands for the occurrence of the corresponding event, and $0$ for the absence.
The collection of EHRs is a matrix $\mathbf{X} \in \{0, 1\}^{N \times M}$ composed by stacking $N$ patient vectors $\{\boldsymbol{x}\}_{1:N}$.
It is challenging to build effective recommendation systems using solely the EHR data, mainly due to the following reasons: 
(1) binary data poses a challenge to effective machine learning: while both 0 and 1 play an equally important role in identifying a proper recommendation, in practice, 0 entries generally do not propagate learning signals;
(2) medical events differ across platforms, and as a result, two binary matrices may share the same shape but have entirely different underlying meanings. 
Our goal is to mine the complete information hidden in binary EHR data effectively, based on which an effective, general-purpose recommendation system can be built.
Formally, \method can be described by
$F(\cdot): \{0, 1\}^M \rightarrow \{0, 1\}^C$,
which accepts an EHR sample as input, and outputs $C$ binary indicators indicating whether or not recommend drug.

\subsection{EHR as a Sample from the Population}\label{sec: EHR_sample}
This paper introduces a statistical method to address binary data. 
Several studies suggest prevalent diseases and medical conditions often share commonalities across populations due to factors like environment and behavior \citep{bache2013adaptable}. 
In terms of epidemiology, temporal and spatial specificity lead to closer medical relatedness within sub-cohorts \citep{sun2012supervised,bache2013adaptable,poulakis2022multi}. 
Based on this, we assume that \textbf{(i)}the outcome of a medical event $E_i$ follows a Bernoulli distribution for the population, and \textbf{(ii)}the EHR dataset is an i.i.d. sample from this population.


To address the issue of signal-stopping 0 entries, we represent them using the Bernoulli mean $1 - P(E_i=1)$, empirically estimated from the EHR dataset, which assumes the same distribution as the population. This transforms binary 0-1 events into continuous representations, helping mitigate the problem of data sparsity. In deep learning models, a value of 0 prevents activation of connection weights, and in graph-based models, zeros do not contribute to node attributes or message passing. 
The Bernoulli mean, ranging from 0 to 1, provides more effective representations of patient-event and event-event relationships, leading to more accurate medication recommendations than binary signals \citep{2015-smokedrink}.

\section{Methodology}

\subsection{Patient-Event Graph Construction}
\label{sec: Patient-Event Graph Construction}

We propose using GNNs to learn the relationships between medical events within each patient and across the patient population. Fig. \ref{fig:main_graph} provides a system overview. For each patient, we construct a graph $\mathcal{G}_n = (\mathcal{V}_n, \mathcal{E}_n)$, where $\mathcal{V}_n$ represents patient-event relationships and $\mathcal{E}_n$ captures event-event connections. Na\"ively initializing $\mathcal{V}_n$ and $\mathcal{E}_n$ from $\mathbf{X}$ could lead to information loss from zero entries, as they hinder signal propagation. 
To this end, we propose an alternative initialization method for graph construction.

\noindent\textbf{Patient-to-patient correlation embedding. }
By assumption\textbf{(i)} in Section \ref{sec: EHR_sample}, the occurrence of an event is subject to a Bernoulli distribution $P(E_j = k) = \rho^k (1-\rho)^k$, where $k \in \{0, 1\}$, and $\rho$ denotes the mean of the distribution.
By assumption\textbf{(ii)}, the event in the EHR data sample is subject to the same distribution $P(\mathbf{X}_{\cdot j} = k) = \rho^k (1-\rho)^k$, where we denote the $\mathbf{X}_{\cdot j}$ denote the event $E_j$ for all patient samples.
Therefore, we propose initializing the vertices by the Bernoulli mean $\rho$. 
While this quantity is not known, we can replace it with a sample estimate:
\begin{equation}
    \tilde \rho := \tilde{P}(X_{\cdot j} = 1) 
 = \frac{\sum_{i=1}^{N} \mathbf{X}_{ij}}{N}, \,\, \tilde{P}(X_{\cdot j} = 0) = 1 - \tilde\rho
\end{equation}

\noindent\textbf{Event-to-event correlation embedding. }
Then, we propose to initialize the edges representing the event-to-event relationship using conditional Bernoulli probability. 
Intuitively, an edge from event $j$ to event $i$ encodes some information \emph{conditional on} the observation of $i$ \citep{2020learningStructure, murphy2023probabilistic}.
From this consideration, we initialize the edge $e_{ij}$ from event $E_j$ to event $E_i$ with:
\begin{equation}
    e_{ij} := P( E_i = 1 | E_j = 1) =  \frac{{P}(E_i=1 \text{ and }  E_j=1)}{{P}(E_j = 1)}
\end{equation}

Since the exact distribution is not known, we hence replace it with a sample estimate. 
That is, the conditional probability is instead estimated by 
$\frac{P(\mathbf{X}_{\cdot i} = 1 \text{ and } \mathbf{X}_{\cdot j} = 1)}{P(\mathbf{X}_{\cdot j} = 1)}$
\noindent
where $P(\mathbf{X}_{\cdot i} = 1 \text{ and } \mathbf{X}_{\cdot j} = 1)$ is the ratio of patients with both events $i$ and $j$ being positive, and $P(\mathbf{X}_{\cdot j} = 1)$ is the ratio of patients with event $j$ being positive.
It is worth noting that the proposed initialization can be seen as a transformation from binary data to continuous values.

\subsection{Patient-to-Event Graph Learning}
\label{sec: Graph Learning}
In Section \ref{sec: Patient-Event Graph Construction}, we mapped each patient sample $\mathbf{X}_{i\cdot}$ to a graph $\mathcal{G}_i = (\mathcal{V}_i,  \mathcal{E}_i)$, with empirical Bernoulli mean initialization. 
This transformation converts the original drug recommendation binary classification problem into a graph classification problem.
Specifically, we first map the EHR data into a set of vertices and edges $ \{0, 1\}^M \stackrel{\text{B.I.}}{\rightarrow} \mathcal{V}_i,  \mathcal{E}_i \in [0, 1]^M $, where B.I. denote Bernoulli Initialization.
A graph learning function is then applied to gain drug recommendation outputs $F(\cdot): \mathcal{V}_i,  \mathcal{E}_i  \rightarrow \{0, 1\}^C$.
Notably, we employ EGraphSage \citep{EgraphSage} as our $F$, where the GNN learns the node and edge embeddings $h_{\nu}, h_{e}$ by an MLP. 
The first layer has an interpretation of nonlinearly transforming the Bernoulli statistics residing on the graph we initialized.
Graph learning aims to discern correlations between events through message aggregation and learn the event (node) attributes.
The message $m_j$ on each edge $e_{ij}$ from node $i$ to node $j$ is:
\begin{equation}
m_j = \text{relu}\left(\mathbf{W}_m \times \left(h_{\nu_j} \oplus h_{e_{ij}}\right)\right)
\end{equation}
where $\oplus$ represents concatenation. $\mathbf{W}_m$ is a learnable weight matrix for message passing, and $\text{relu}$ is a non-linear activation function.

\noindent\textbf{Message aggregation:} 
Afterward, we aggregate messages $m_j$ across edges for each node:
\begin{equation}
\text{AGGREGATE}_i = \frac{1}{|\mathcal{N}(i)|} \sum_{j \in \mathcal{N}(i)} m_j
\end{equation}
where $ \mathcal{N}(i) $ represents the set of neighbors.

\noindent\textbf{Update node features:} 
We update and normalize the node feature $h_{\nu_i}$ for each node $\nu_i$ using:
\begin{equation}
h_{\nu_i} = \text{activation}\left(\mathbf{W}_u \times \left(h_{\nu_i} \oplus \text{AGGREGATE}_i\right)\right)
\end{equation}
where $h_{\nu_i}' = \frac{h_{\nu_i}'}{\|h_{\nu_i}'\|_2}$ is the updated feature of node $i$, and $\mathbf{W}_u$ is learnable weight matrix for updating, $\|.\|_2$ denotes the L2 norm.

\begin{algorithm}[t]
\caption{
Pseudocode of patient-event graph
}
\label{alg:generate_graph}
\lstset{caption={}}
\begin{lstlisting}[language=python]
def generate_patient_graph(EHR_Matrix):
    # EHR_Matrix: Rows represent patients, and columns represent medical events
    # Calculate proportions of each event based on all patients
    event_proportions = mean(EHR_Matrix, axis=0)
    # Calculate posterior probabilities between each pair of events for the pair of event i and event j:
    posterior probability(event i, event j) = $\frac{{\text{Proportion}}(\text{event i} = 1 \text{ and }  \text{event j} = 1)}{{\text{Proportion}}( \text{event j} = 1)}$ = $\frac{\text{number of} (\text{event i} == 1 \;\text{and}\; \text{event j} == 1)}{ \text{number of all patients}} / \frac{\text{number of} (\text{event j} == 1)}{\text{number of all patients}}$
    For each patient, generate a graph:
        # Determine node embeddings based on the patient's medical event data
        node_embeddings = event_proportion if input value == 1
        node_embeddings = 1-event_proportion if input value == 0
        # Determine edge weights by the posterior probabilities between each pair of events
        edge_weights of edge from node i to node j = 
            posterior probability of (i, j)

\end{lstlisting}
\end{algorithm}
\subsection{Medication Recommendation}
\label{sec: Drug Recommendation Inference}

After $K$ iterations, the last layer updated node feature $(h'_{\nu}, \forall{\nu \in \mathcal{V}})$ serves as the output of our graph embedding learning part.
After obtaining the node embeddings $ h'_{\nu} $ for each graph, we concatenate these embeddings to form a single feature vector for each graph. This is represented as:
\begin{equation}
\mathbf{z}_{\mathcal{G}_i} = \oplus_{(\nu \in \mathcal{V})} h'_{\nu},
\end{equation}
where $\mathbf{z}_{\mathcal{G}_i}$ is the concatenated feature vector for graph $\mathcal{G}_i$ and $\oplus$ denotes the concatenation operation over all nodes (i.e., medical events).
The concatenated feature vectors $\mathbf{z}_{\mathcal{G}_i}$ for each graph are then passed through a linear projection for classification. The linear layer has $Y$ independent output dimensions, each corresponding to a binary classification task. Formally, for each output dimension $y$:
\begin{equation}
\mathbf{o}_y = \text{sigmoid}\left(\text{Linear}\left(\mathbf{z}_{\mathcal{G}_i}\right)\right)
\end{equation}
\begin{equation}
\mathbf{r}_y = 
\begin{cases} 
1 & \text{if } \mathbf{o}_y > 0.5 \\
0 & \text{otherwise}
\end{cases}
\end{equation}

Upon processing graph feature vectors $\mathbf{z}_{\mathcal{G}_i}$, the system yields $Y$ independent outputs. Each of these outputs corresponds explicitly to the decision to recommend drugs. 
The decision is binary: either the medication is recommended (represented by a value of 1) or not (denoted by 0).

\section{Experiments}\label{sec:experiment}
\subsection{Databases}\label{subsec:database}
We evaluated \method on two large-scale datasets: the Medical Information Mart for Intensive Care (MIMIC-III) dataset and the Antimicrobial Resistance in Urinary Tract Infections (AMR-UTI).
Both databases have been extensively utilized for benchmarking and are available on PhysioNet. 

\noindent - \textbf{MIMIC-III} represents a vast repository of de-identified health data from intensive care unit (ICU) patients at the Beth Israel Deaconess Medical Center in Boston, Massachusetts \citep{Johnson2016MIMICIIIAF}. 
It contains a total of 46520 patients from 2001 to 2012. 

\noindent - \textbf{AMR-UTI} dataset contains EHR information collected from 51878 patients from 2007 to 2016, with urinary tract infections (UTI) treated at Massachusetts General Hospital and Brigham \& Women's Hospital in Boston, MA, USA \citep{Michael2020AMR-UTI}.
This dataset provided a series of binary indicators for whether a patient was undergoing a specific medical event.
There are four common antibiotics (medication labels): nitrofurantoin (NIT), TMP-SMX (SXT), ciprofloxacin (CIP), or levofloxacin (LVX).

Table \ref{tb1:database} shows some statistics on two databases. 
The sparsity of a matrix can be calculated using the formula: $Sparsity = 1 - \frac{\text{Number of Non-Zero Elements}}{\text{Total Number of Elements}}$. 
The high degree of sparsity observed in the feature dimensions for both the MIMIC-III and AMR-URI datasets may constitute a significant impediment to the efficacy of the machine learning models.

\begin{table}[ht]
\centering
\caption{Datasets description}
\begin{tabular}{l|cc}
\toprule
Dataset                                & MIMIC              & AMR-URI \\ \midrule
\#Patients & 46250 &    51878       \\
Avg. event sequence length       & 3388 &    692       \\
\#Drug candidate        & 131 &    4      \\
The Sparsity of event seq.               & 
       0.9946 & 0.9309              \\
Sparsity of label   & 0.8534       &   0.5423          \\
\bottomrule
\end{tabular}
\label{tb1:database}
\end{table}

\noindent \textbf{Data preprocessing.}
For MIMIC-III, we referred to the data processing released by \citep{safedrug, wwwcognet} for fairness.
Only the patients with at least 2 visits are incorporated.
The medications were selected and retained based on their frequency of occurrence (the top 300).
For AMR-URI, we first excluded the basic demographic information including age and ethnicity. 
The observations that do not have any health event or drug recommendation were removed.
For each observation, a feature was constructed from its EHR as a binary indicator for whether the patient is undergoing a particular medical event within a specified time window.
After the above preprocessing, we divided both datasets into training, validation and testing by the ratio of 3/5, 1/5, and 1/5. 
More detailed description of data processing can be found in Appendix \ref{appendix:Preprocess}.

\begin{table*}[t]
	\caption{Performance Comparison on MIMIC-III and AMR-UTI. 
 Best results are highlighted in \textbf{bold fonts}, second-best results are \underline{underscored}. 
 The results (-) denote models that do not converge.
 }
	\centering
	\label{tab:mainresults}
 \resizebox{0.98\linewidth}{!}{%
	\begin{tabular}{l|lccccc|cccccl}
		\toprule
Dataset              & \multicolumn{6}{c}{MIMIC}                     &  & \multicolumn{5}{c}{AMR-UTI}         \\ 
Baseline              & Methods & Jaccard & F1 & PRAUC & AUROC & \#Drug &  & Jaccard & F1 & PRAUC & AUROC & \#Drug \\ 

		\midrule
LR                    & Stat    & 0.4865                      & 0.6434                 & 0.7509                    & 0.9180                    & 16.1773                 &  & 0.3803       & 0.5018  & 0.4624     &  0.5057    & 1.4869    \\
ECC                   & Stat    & 0.4996                     & 0.6569                 & 0.6844                    & 0.9098                    & 18.0722                 &  & \underline{0.6080}       &  \underline{0.6665}  &  \underline{0.6914}     &\underline{0.7273}    & 1.7040    \\
RETAIN         & RNN     & 0.4877                     & 0.6481                 & 0.7556                   & 0.9234                    & 20.4051                  &  & -      & -  & -   & -     & -    \\
LEAP           & RNN    & 0.4521                      & 0.6138               & 0.6549                  & 0.8927                    & 18.7138                &  & -     & -  & -   & -     & -    \\
DMNC            & RNN     & 0.4864                    & 0.6529                 & 0.7580                  & 0.9157                    & 20.0000               &  & -       & -  & -     & -     & -   \\
SafeDrug       & RNN     &  0.5213                      & 0.6768                 & 0.7647                    & 0.9219                    & 19.9178                  &  & -      & -  & -     & -     & -    \\
GAMENet         & Graph    & 0.5067                      & 0.6626                 & 0.7631                    & \underline{0.9237}                    & 27.2145                  &  & 0.6024       & 0.6433  & 0.6913     & 0.7174     & 1.7106    \\
COGNet         & Graph    & 0.5336                      & 0.6869                 & 0.7739                    & 0.9218                    & 28.0903                  &  & 0.5287    & 0.6359  & 0.5717  &0.6281         & 1.6331   \\
MT-GIN           & Graph & \underline{0.5401}       & \underline{0.7781}  & \underline{0.7812}     & 0.9115     & 16.1233    &  & 0.5137       & 0.6124  & 0.6779     & 0.6428     & 1.4991    \\
		\midrule
		\method & Graph   & \textbf{0.5887}                      & \textbf{0.8459}                 & \textbf{0.8442}                    & \textbf{0.9632}                    & 15.7129                  &  & \textbf{0.6116}       & \textbf{0.6753}  & \textbf{0.7071}     & \textbf{0.7401}     & 1.5973    \\ 		\bottomrule
	\end{tabular}
 }
\end{table*}

\subsection{Baselines}\label{subsec:baseline}
\noindent - \textbf{Statistics-based methods:} We use standard Logistic Regression (LR) and Ensemble Classifier Chain (ECC) \citep{ecc} as baselines.
\noindent - \textbf{RNN-based methods:} RETAIN \citep{retain} applies two-level attention with gating for key clinical variables. LEAP \citep{2017-LEAP} uses an LSTM-based model with reinforcement learning to avoid adverse drug combinations. DMNC \citep{dmnc} introduces a memory-augmented network to enhance patient encoding.
\noindent - \textbf{GNN-based methods:} GAMENet \citep{gamenet} incorporates Drug-Drug Interaction (DDI) knowledge using graph convolutional networks. SafeDrug \citep{safedrug} combines the drug molecular graph with DDI to predict safe medication combinations. Using a Conditional Generation Net, COGNet \citep{wwwcognet} mines relationships between historical and current diagnoses. MI-GIN \citep{2023-adma} directly models EHR data using a graph isomorphism neural network for medication recommendation.
\noindent - \textbf{Evaluation:} We evaluated models using the Jaccard Similarity Score, Average F1, AUROC, Precision-Recall AUC, and average predicted number of drugs (\#Drug), averaging results overall for patients. Precision and Recall were also used to assess recommendation performance. Parameter settings are detailed in Appendix \ref{appendix:ModelParameters}.

\begin{figure*}[t]
    \centering
    \includegraphics[width=2\columnwidth]{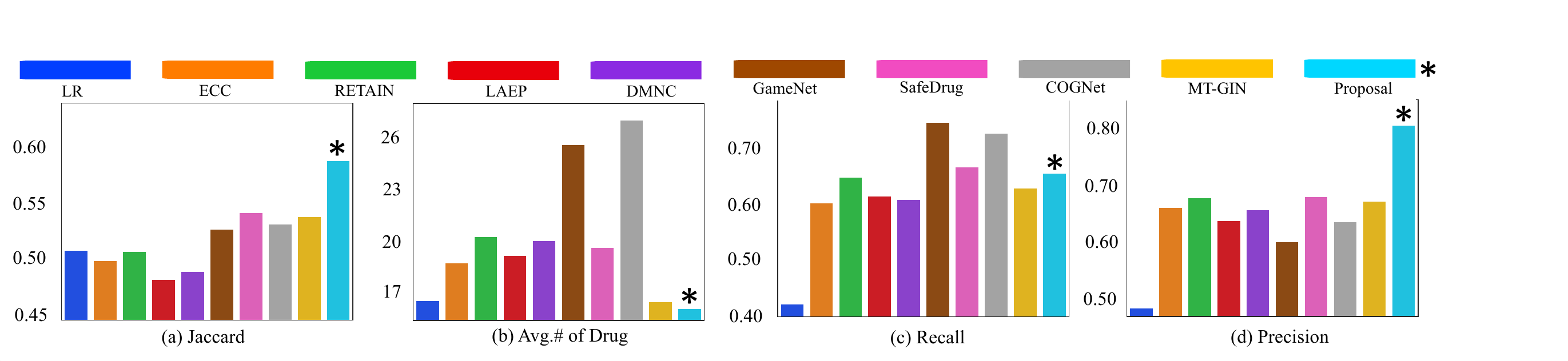}
  \caption{
  Comparison between \method and baselines using four widely adopted metrics (Jaccard, Avg of Drug, Recall, Precision) on MIMIC III for medication recommendation analysis.
 }
 
  \label{fig: Histogram}
\end{figure*}

 \begin{table*}[t]
    \caption{
    An ablation study on embeddings is conducted to validate the contributions of the proposed embedding approach. 
    The best-performings are highlighted in \textbf{bold}, while the second-best results are \underline{underscored}.}
	\centering
	\label{table_3}
 \resizebox{\linewidth}{!}{%
	\begin{tabular}{l|ccccc|llllll}
		\toprule
Dataset              & \multicolumn{5}{c}{MIMIC}                     &  & \multicolumn{5}{c}{AMR-UTI}         \\ 
Metrics               & Jaccard & F1 & PRAUC & AUROC & \#Drug &  & Jaccard & F1 & PRAUC & AUROC & \#Drug \\

		\midrule
		BM $w/o$ Post  & 0.5482             & \underline{0.8275}            & 0.151                    & \underline{0.9542}                  & 15.1754                  &  &\underline{0.6109}       & \underline{0.6750}  & \underline{0.7066}     & \underline{0.7378}     & 1.5966    \\
  		LLR $w/$ Post  & 0.5398              & 0.8234            & \underline{0.8104}                    & 0.9527                    & 14.8922                  &  & 0.6091              & 0.6748            & 0.7052                    & 0.7322                    & 1.5990   \\
		LLR $w/o$ Post  & 0.4367              & 0.7717            & 0.7265                    & 0.242                    & 12.7554                 &  & 0.6089              & 0.6732            & 0.7048                    & 0.7318                    & 1.5964 \\
		TE $w/$ Post  & 0.3904              & 0.7475          & 0.6947                    & 0.9116                    & 11.7528                 & & 0.0186             & 0.3580          & 0.4961                    & 0.5387                   &0.0172  \\
		TE $w/o$ Post  & 0.3904              & 0.7475          & 0.6947                    & 0.9116                    & 11.7528               &  & 0.0001              & 0.3507          & 0.4595                    & 0.5000                    & 0.0001     \\
            BM $w/$ RE  &\underline{0.5692} & 0.8261 &0.8230 &0.9428 &15.8093   &  & 0.6019 &0.6579  &0.6881 &0.7218 &1.6220   \\
            
		RN $w/$ Post  & 0.2941 & 0.6674 & 0.4902 &0.8174 &24.8895  & & 0.5974 &0.3554  &0.4685 &0.4776 &2.9293    \\
    
		RN $w/$ RE  &0.2932   &0.6640   &0.4899            &0.8075 &24.8788 &  &0.5963 &0.3530 &0.4601 & 0.4744 &2.9275 \\
  \midrule
  
   MLP backbone   & 0.3926 & 0.7484 & 0.6871 & 0.8948 & 12.0562 & & 0.5667 & 0.6685 & 0.6945 & 0.7137 & 1.4853  \\
            
		LSTM backbone& 0.3868 & 0.7454 & 0.6940 & 0.9102 & 11.4403 &  &   0.5352 & 0.6491 & 0.6810 & 0.7217 & 1.3640   \\
    
		Transformer backbone& 0.3907 & 0.7475 & 0.6867 & 0.8996 & 11.9148 &  &0.5100 & 0.3512 & 0.4662 & 0.5088 & 1.3470 \\
            \midrule
            \textbf{Ours} (\textbf{BM $w/$ Post})  & \textbf{0.5887}                      & \textbf{0.8459}                 & \textbf{0.8442}                    & \textbf{0.9632}                    & 15.7129                  &  & \textbf{0.6116}       &\textbf{0.6753}  & \textbf{0.7071}     & \textbf{0.7401}     & 1.5973 \\   
  \bottomrule
	\end{tabular}
 }
\end{table*}

\subsection{Ablation Study}\label{subsec:ablations}

We evaluated different node/edge modeling methods, with our key contribution being the transformation of binary medical events into continuous Bernoulli means. We compared \method with two classic encoding methods, Log Likelihood Ratio (LLR) and Target Encoding (TE) \citep{2018-TargetEncoding} (details in Appendix \ref{app:metrics}). Additionally, we performed ablation studies using random node/edge initialization and assessed \method without node or edge construction.
We evaluated several variations of our method. BM $w/$ Post is our proposed approach. BM $w/o$ Post uses simple co-occurrence for edge embeddings. LLR $w/$ Post replaces Bernoulli means with LLR for node embeddings, while LLR $w/o$ Post uses LLR for both. TE $w/$ Post and TE $w/o$ Post follow a similar pattern with Target Encoding. BM $w/$ RE uses random edge initialization, and RN $w/$ Post and RN $w/$ RE replace node or both node/edge embeddings with random initialization. We also conducted ablation studies comparing GNN event-event modeling with MLP, LSTM, and Transformer. Details are in Appendix \ref{appendix:ImplementationDetails}.

\section{Results}

\subsection{Comparison Against Baselines}

\noindent \textbf{Main Results.}
Table~\ref{tab:mainresults} presents the main results for medication recommendation on two datasets. Our proposed \method outperforms all baselines across all metrics (Jaccard, F1, PRAUC, AUROC), particularly excelling on the MIMIC-III dataset, even against models using additional data and visit modeling. Statistical models like LR and ECC surpass early RNN-based methods (LEAP, DMNC, RETAIN), highlighting the strong performance of ECC. RNN models struggle with missing or inconsistent historical data, while graph-based methods perform better by modeling event-event correlations, showing the effectiveness of modeling event-event correlations and structure information. \method, based solely on binary events, could improve by incorporating external EHR data like DDI.

\noindent \textbf{Visit-Included Results.}
The results presented in Table~\ref{tab:mainresults} exclude information derived from patient visits. To provide a more comprehensive assessment incorporating visit data, we present an average recommendation performance across all patient visits in Table \ref{table_2}.
Specifically, some of the baseline methods rely on sequential learning.
In these approaches, every medication is allocated a recommendation probability upon generation in each visit.
We first average their performance over visits for each patient and then further average these results across all patients.
In this context, \method shows an effective and robust performance and outperforms all baselines.

\noindent \textbf{Medication Recommendation Analysis.}
Fig. \ref{fig: Histogram} extensively compares Precision and Recall metrics for medication recommendation.
We employ several evaluation metrics to compare \method against existing methods.
The False Positive (FP) issue arises when the system suggests unnecessary drugs, leading to high recall but low precision. 
Conversely, False Negatives (FN) occur when beneficial drugs are missed, resulting in low recall but potentially high precision.
As shown in Fig. \ref{fig: Histogram}, we provide 15.7129 recommendations on average, and the ground truth for MIMIC-III is 19.2268.
This reflects that our method can recommend fewer but more precise drugs, maintaining competitive recall.

\begin{table}
	\caption{Comparison when visiting information is included on MIMIC-III. Though \method is not explicitly designed to incorporate secondary information, it outperforms existing visiting-aware methods when evaluated on the same benchmarks. \textbf{Bold}: best; \underline{underscored}: second-best.}
	\centering
	\label{table_2}
    \resizebox{\columnwidth}{!}{
	\begin{tabular}{l|ccccc}
		\toprule
		Model    & Jaccard & F1 & PRAUC & AUROC   & \#Drug \\
		\midrule
		LR       & 0.4844 & 0.6483 & 0.7492 & 0.9029 & 16.1221 \\
            ECC       & 0.5003 & 0.6590 & 0.6802 & 0.9214 & 18.4127 \\
		RETAIN   & 0.4876 & 0.6484 & 0.7594 & 0.9225 & 19.8297 \\
		LEAP     & 0.4487 & 0.6119 & 0.6480 & 0.8927 & 19.1095 \\
		DMNC     & 0.4834  & 0.6492 & 0.7569 & 0.9237 & 20.1087 \\
		GAMENet  &  0.5084  & 0.6648 & \ul{0.7670} & 0.9235 & 26.1983 \\
		SafeDrug & \ul{0.5276}   & 0.6692  & 0.7649 & 0.9218 & 19.8847 \\
            COGNet & 0.5127  & 0.6850  & 0.7739  &  \ul{0.9266}  & 27.3277 \\ 
            MT-GIN & 0.5273   & \ul{0.7527}  & 0.7499 & 0.9188 & 15.1233 \\
		\midrule
  \textbf{Proposed}       & \textbf{0.5902}  & \textbf{0.8465} &  \textbf{0.8448} & \textbf{0.9635} & 15.6944 \\
		\bottomrule
	\end{tabular}
 }
\end{table}

\subsection{Ablation Study}

Table \ref{table_3} shows that combining Bernoulli mean node embedding with posterior probability edge embedding yields the best results. 
If we compare the contributions of node (Bernoulli mean) and edge (posterior probability) embeddings to the graph-based drug recommendation results, we find that node embedding has a relatively more substantial impact.
The random node destroys the recommendation performance even when considering posterior probabilities, resulting in a low Jaccard score of 0.29.
However, using the Bernoulli mean for edge initialization yields a much higher Jaccard score of 0.5692.
This could be because node embedding directly represents the original EHR data, while edges are inferred based on the original data to model the relationship between different events. 
Therefore, edge embeddings merely aid in GNN computations. 
The results show that even without initializing edge embeddings with posterior probabilities, GNN can still capture event-event correlations, though with a performance decrease. 
For model ablation, GNN outperforms all baselines.

\begin{figure}[t]
    \centering
\includegraphics[width=0.9\columnwidth]{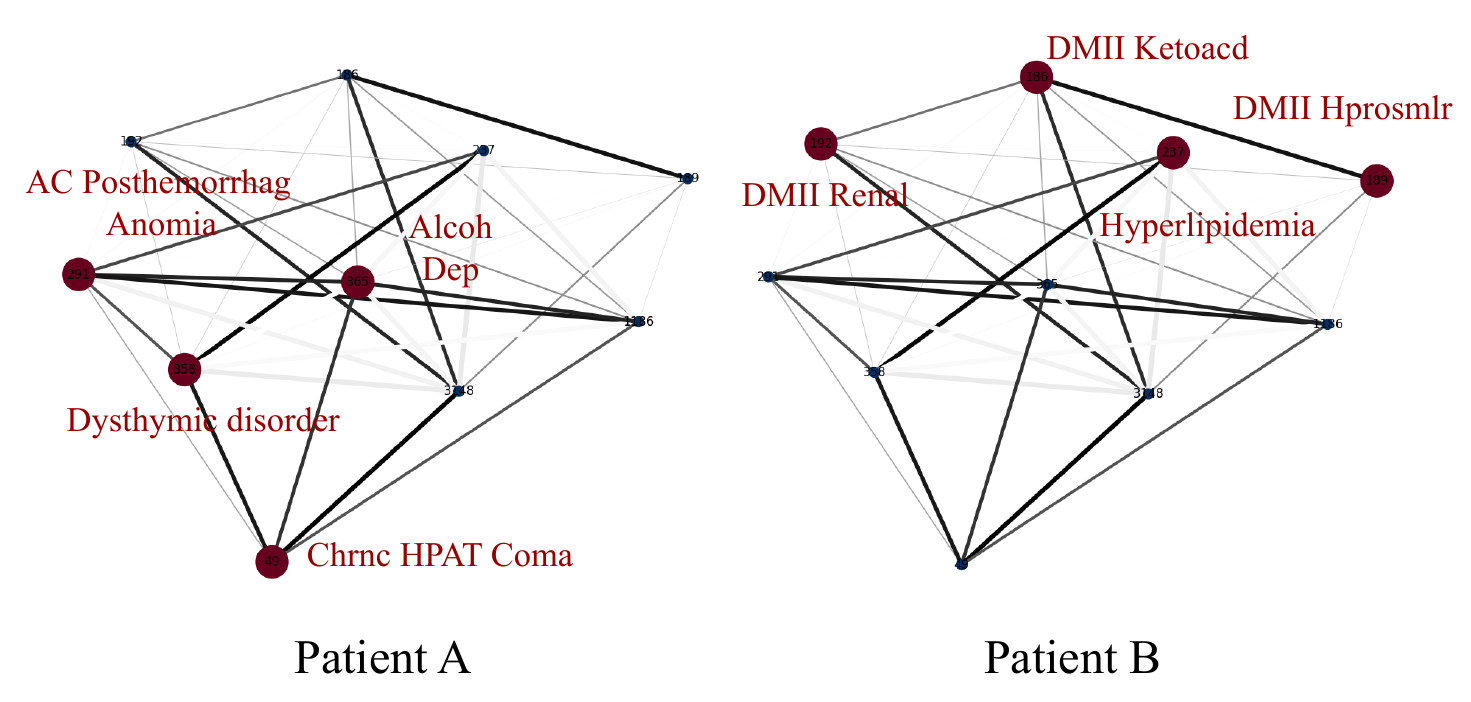}
\caption{Feature visualization:
Each node denotes an event, and edges represent event-event correlations.
A thicker edge shows a stronger event-event association, highlighting higher-value nodes.
}
\label{fig:GNN'soutput}
\end{figure}

\subsection{Visualization of Recommendations}

Fig. \ref{fig:GNN'soutput} shows GNN-learned representations for two patients on a subset of EHR events. 
We highlight the top nodes with the highest feature values by TOP-K, revealing that GNN captures distinct representations for each patient. 
For example, the "Hyperlipidemia" node is more prominent in Patient B than in Patient A, indicating greater relevance. 
GNN also learns key feature connections, with certain event series highlighted in Patient B. 
This demonstrates the ability of GNN to differentiate between patients and provide valuable insights for downstream tasks. 

\section{Conclusion}\label{sec:conclusion}

This paper introduces \method, a novel approach that transforms sparse 0/1 binary event sequences in EHRs into continuous Bernoulli means. Viewing EHR data as samples from a population, we modeled GNN nodes and edges using the Bernoulli distribution. 
Our method outperformed all baselines, including those using more extensive data like visit records and drug-drug interactions. 
Ablation studies confirmed the importance of Bernoulli transformation, with node-based Bernoulli means contributing significantly to performance, even with random edges. 
This work presents a promising direction for handling sparse binary sequences in EHRs, potentially improving various healthcare prediction tasks.

\bibliography{main}

\clearpage
\appendix

\onecolumn

\begin{adjustwidth}{1cm}{1cm}

\begin{center}
{\LARGE \textbf{BernGraph: Probabilistic Graph Neural Networks for EHR-based Medication Recommendations\\ $\;$ \\ ————Appendix————}}
\end{center}

\begin{center}
  \hypersetup{hidelinks}
  \tableofcontents
  \noindent\hrulefill
\end{center}

\end{adjustwidth}

\twocolumn



\section{Preprocessing}
\label{appendix:Preprocess}

\subsection{Data Profile}

\textbf{Data Sources}:
The data is sourced from the MIMIC dataset, a publicly available dataset developed by the MIT Lab for Computational Physiology, comprising de-identified health data from over 40,000 critical care patients. The primary datasets in focus for preprocessing include:

\begin{itemize}

   \item Medications (`PRESCRIPTIONS.csv`): Contains records of medications prescribed to patients during their ICU stays.
   
   \item Diagnoses (`DIAGNOSES ICD.csv`): Lists diagnosis codes associated with each hospital admission.
   
   \item Procedures (`PROCEDURES ICD.csv`): Catalog procedures that patients underwent during their hospital visits.
\end{itemize}
\textbf{Data Structure}:
The data is structured in tabular format. Each table contains unique identifiers for patients (`SUBJECT ID`), their specific hospital admissions (`HADM ID`), and other clinical details pertinent to the dataset in question.

\subsection{Medication Data Preprocessing}

\textbf{Data Loading and Initial Processing}:

\begin{itemize}
   \item Load the Data: The medication data from `PRESCRIPTIONS.csv` is loaded into a data frame.
   
   \item Filter Columns: Only essential columns, namely `pid`, `adm id`, `date`, and `NDC`, are retained.
   
   \item Data Cleaning: Entries with 'NDC' equal to '0' are dropped, and any missing values are filled using the forward-fill method.
   
   \item Data Transformation: The `STARTDATE` field is converted to a DateTime format for easier manipulation and analysis.
   
   \item Sorting and Deduplication: The dataset is sorted by multiple columns, including `SUBJECT ID`, `HADM ID`, and `STARTDATE`, and any duplicate entries are removed.
\end{itemize}

\textbf{Medication Code Mapping}:

\begin{itemize}
   \item Mapping NDC to RXCUI: The `NDC` codes are mapped to the `RXCUI` identifiers using a separate mapping file.
   
   \item Mapping RXCUI to ATC4: The `RXCUI` identifiers are further mapped to the `ATC4` codes using another mapping file. The ATC classification system is crucial for grouping drugs into different classes based on their therapeutic use.
\end{itemize}

\subsection{Diagnosis Data Preprocessing}

The diagnosis data from `DIAGNOSES ICD.csv` is processed using the `diag process` function, which involves:
\begin{itemize}
   \item Load the Data: Diagnosis data is loaded into a data frame.
   
   \item Filter Relevant Columns: Only essential columns, which might include diagnosis codes and patient identifiers, are retained.
   
   \item Handle Missing or Erroneous Entries: Any missing values or errors in the dataset are addressed, ensuring data integrity.
   
   \item Sorting and Deduplication: The dataset is sorted based on relevant columns, and duplicate entries are removed to ensure each record is unique.
\end{itemize}
\subsection{Procedure Data Preprocessing}

Similar to the diagnosis data, the procedure data from `PROCEDURES ICD.csv`:
\begin{itemize}
   \item Load the Data: Procedure data is loaded into a DataFrame.
   
   \item Filter and Clean: Only relevant columns are retained, and any erroneous or missing entries are addressed.
   
   \item Sorting and Deduplication: The data is organized by relevant columns, and any duplicate records are removed.
\end{itemize}
\subsection{Data Integration}

\textbf{Combining Process}:
The datasets, once cleaned and preprocessed, need to be integrated for comprehensive analysis:
\begin{itemize}
    \item Data Merging: The medication, diagnosis, and procedure data are merged based on common identifiers like `SUBJECT ID` and `HADM ID`.
   
    \item Final Cleaning: Any discrepancies resulting from the merge, such as missing values or duplicates, are addressed.
\end{itemize}

\section{Model Parameters}
\label{appendix:ModelParameters}

GNN Model (EGraphSage)
The Graph Neural Network model is implemented using the EGraphSage class. The configuration parameters for the model are in the Table \ref{appendix:parameters}

\begin{table}
\centering
\caption{Datasets description}
\label{appendix:parameters}
\begin{tabular}{l|ll}
\hline
Dataset                                & GNN              & MLP \\ \hline
Input Dimension        & 1 &    \#nodes       \\
Hidden Layer Dimension  & 128 &    64       \\
Output Dimension        & 1 &    1       \\
Edge Channels       & 1 &    NaN       \\
Activation Function       & ReLU &    ReLU       \\
Aggregation Method        & Mean &    NaN       \\
\hline
\end{tabular}
\end{table}

\section{Metrics}\label{app:metrics}
Target encoding is a popular technique in machine learning for encoding categorical features, where each category is replaced with its corresponding mean target value. This can be mathematically expressed as:
\[
\text{Target Encoding}(c) = \frac{\sum_{i=1}^{N} y_i \cdot \delta(c, c_i)}{\sum_{i=1}^{N} \delta(c, c_i)}
\]
where \(c\) represents the category to be encoded, \(y_i\) is the target value for the \(i\)-th sample, \(c_i\) is the category of the \(i\)-th sample, and \(\delta(c, c_i)\) is the Kronecker delta function that equals 1 when \(c\) equals \(c_i\) and 0 otherwise.

\textit{Log Likelihood Ratio (LLR)}: The likelihood ratio is defined as the ratio of two key values: 
The maximum value of the likelihood function is calculated within the subspace defined by the hypothesis. 
And the maximum value of the likelihood function, calculated across the entire parameter space \citep{dunning1994accurate}.  

To compute the score, let $k_{11}$ be the number of times the events occurred together, let $k_{12}$ and $k_{21}$ be the number of times each has occurred without the other, and $k_{22}$ be the number of times something has been observed that was neither of these events. The LLR score is given as the following:
\[
LLR = 2 \sum(k) \left(H(k) - H(\text{rowSums}(k)) - H(\text{colSums}(k))\right)
\]
where $H$ is Shannon's entropy, computed as the sum of:
\[
H = \frac{k_{ij}}{\sum(k)} \log \left(\frac{k_{ij}}{\sum(k)}\right).
\]

In our case, when calculating the similarity measure between a medical event and the recommendation of a specific drug, we calculated the LLR value by the counts that co-occurred between them.
The average LLR for each medical event with all recommended drugs was calculated as the properties of the medical event node
\citep{dunning1994accurate}.

\section{Implementation Details}
\label{appendix:ImplementationDetails}

The models mentioned were implemented by PyTorch 2.0.1 based on Python 3.8.8, 
All experiments are conducted on an Intel Core i9-10980XE machine with 125G RAM and an NVIDIA GeForce RTX 3090. 
For the experiments on both datasets, we chose the optimal hyperparameters based on the validation set.
Models were trained on Adam optimizer with learning rate $1\times10^{-4}$ for 200 epochs. 
The random seed was fixed as 0 for PyTorch to ensure the reproducibility of the models.

For a fair comparison, we employed bootstrapping sampling instead of cross-validation in the testing phase.
Specifically, in each evaluation round, we randomly sampled $80\%$ of the data from the test set.
We repeated this process 10 times to obtain the final results. 
These results from 10 rounds were then used to calculate both the mean and standard deviation.
The means are reported in Table \ref{tab:mainresults}.
The standard deviations are all under 0.05.




\end{document}